\documentclass{article}
\PassOptionsToPackage{numbers, compress}{natbib}
\usepackage[sort, numbers]{natbib}

\usepackage[nonatbib, final]{neurips_2024}

\usepackage[utf8]{inputenc} 
\usepackage[T1]{fontenc}    
\usepackage{hyperref}       
\usepackage{url}            
\usepackage{booktabs}       
\usepackage{amsfonts}       
\usepackage{nicefrac}       
\usepackage{microtype}      
\usepackage{xcolor}         
\usepackage{multirow}
\usepackage{booktabs}
\usepackage{tcolorbox}
\usepackage{graphicx} 
\usepackage{lscape}
\usepackage{enumitem}
\usepackage{svg}

\newtcbox{\blue}[1][]{on line,boxsep=2pt,left=0pt,right=0pt,top=0pt,bottom=0pt,colframe=white,colback=blue!30!white,#1}
\newtcbox{\red}[1][]{on line,boxsep=2pt,left=0pt,right=0pt,top=0pt,bottom=0pt,colframe=white,colback=red!30!white,#1}
\newtcbox{\yellow}[1][]{on line,boxsep=2pt,left=0pt,right=0pt,top=0pt,bottom=0pt,colframe=white,colback=yellow!30!white,#1}
\newtcbox{\green}[1][]{on line,boxsep=2pt,left=0pt,right=0pt,top=0pt,bottom=0pt,colframe=white,colback=green!30!white,#1}
\newtcbox{\orange}[1][]{on line,boxsep=2pt,left=0pt,right=0pt,top=0pt,bottom=0pt,colframe=white,colback=orange!30!white,#1}
\title{What does the Failure to Reason with ``Respectively'' in Zero/Few-Shot Settings Tell Us about Language Models?}

\title{Linq-Embed-Mistral Technical Report}

\author{%
  Chanyeol       Choi, Junseong Kim, Seolhwa Lee, Jihoon Kwon, Sangmo Gu, \\ \textbf{Yejin Kim, Minkyung Cho, Jy-yong  Sohn} \\   \\
  Linq\\ \\
  \texttt{jacob.choi@getlinq.com} \\
}

\begin{document}

\maketitle

\begin{abstract} 
This report explores the enhancement of text retrieval performance using advanced data refinement techniques. We develop Linq-Embed-Mistral\footnote{\url{https://huggingface.co/Linq-AI-Research/Linq-Embed-Mistral}} by building on the E5-mistral and Mistral-7B-v0.1 models, focusing on sophisticated data crafting, data filtering, and negative mining methods, which are highly tailored to each task, applied to both existing benchmark dataset and highly tailored synthetic dataset generated via large language models (LLMs). Linq-Embed-Mistral excels in the MTEB benchmarks (as of May 29, 2024), achieving an average score of 68.2 across 56 datasets, and ranks 1st among all models for retrieval tasks on the MTEB leaderboard with a performance score of 60.2. This performance underscores its superior capability in enhancing search precision and reliability. Our contributions include advanced data refinement methods that significantly improve model performance on benchmark and synthetic datasets, techniques for homogeneous task ordering and mixed task fine-tuning to enhance model generalization and stability, and a streamlined evaluation process using 4-bit precision and a light retrieval evaluation set, which accelerates validation without sacrificing accuracy.
\end{abstract}

\section{Introduction}
In recent years, the convergence of large language models (LLMs) and information retrieval (IR) has garnered significant attention \cite{zhu2023large}. Especially, effective text retrieval is pivotal in integrating LLMs and IR systems as it greatly improves the system's capacity. Enhancing the text retrieval aspect is also crucial for frameworks like Retrieval-Augmented Generation (RAG), which incorporate current external information to overcome the static nature of LLMs, thus delivering dependable and dynamic answers to user inquiries \cite{huang2024survey}. This report explores extensive experiments focused on \textbf{improving text retrieval using advanced data refinement methods}, including sophisticated data crafting, data filtering, and negative mining techniques. These methods are applied to both (1) existing benchmark dataset, and (2) highly tailored synthetic dataset generated via LLMs. Recent studies highlight the efficacy of LLMs in generating synthetic data, primarily for enhancing human-labeled datasets or improving performance \cite{dai2022promptagator, jeronymo2023inpars, wang2023improving, muennighoff2024generative, lee2024gecko}. This motivates us to investigate a critical question: 
\begin{itemize}
\item Can we rely on LLM-generated data to improve retrieval performances? If not, how can we enhance its quality for this specific task?
\end{itemize}

We employ advanced methods such as data crafting with extensive prompt engineering, data filtering, and negative mining guided by teacher models, which are highly tailored to each task, to improve the quality of the synthetic data generated by LLM. Our efforts aim to create high-quality triplet datasets (query, positive example, negative example), significantly improving text retrieval performances.

\subsection{Research Highlights}
Similar to the SFR \cite{SFRAIResearch2024}, our Linq-Embed-Mistral represents a notable progression in text-embedding models, leveraging the robust bases of E5-Mistral \cite{wang2023improving} and Mistral-7B-v0.1 \cite{jiang2023mistral}.

The key experimental points are: 
\begin{itemize}
    \item Linq-Embed-Mistral performs well in the MTEB benchmarks, with an average score of 68.2 across 56 datasets. This places it 1st among publicly accessible models listed on the MTEB leaderboard and 3rd overall.
    \item The model shows a significant enhancement in the retrieval performance, ranking 1st among all models listed on the MTEB leaderboard with a performance score of 60.2.
    \begin{itemize}
        \item Within the Mistral Model Series, a suite of models based on the foundational Mistral architecture, SFR enhances E5-Mistral by adding a specially curated dataset of MTEB tasks. In contrast, our approach focuses solely on \textbf{creating and integrating more sophisticated synthetic datasets}. This has increased our model's score from 56.9 for E5-Mistral and 59.0 for SFR to an 60.2.
    \end{itemize}
\end{itemize}

Our contribution points are as follows:
\begin{enumerate}
    \item Our proposed Data Refinement Methods, which include sophisticated data crafting, filtering, and negative mining, significantly enhance the model's ability to identify misleading documents. By improving the quality of the benchmark dataset and addressing issues in the synthetic dataset generated by GPT-4, these methods ensure more accurate and reliable results.
    \item We propose Homogeneous Task Ordering and Mixed Task Fine-tuning, which enhance the model performance by promoting better generalization and training stability, especially when mixed task fine-tuning is limited to within 20 steps. Here, homogeneous task ordering provides precise insights into task ordering effects, whereas Mixed Task Fine-tuning mitigates the catastrophic forgetting.
    \item We design Streamlined Evaluation,  which uses 4-bit precision and a light retrieval evaluation set. This speeds up the process of validation, where our streamlined evaluation has negligible performance differences, compared with the full-scale evaluation. Our design allows a single GPU to evaluate one checkpoint in approximately 5 hours, with retrieval tasks specifically taking around 4 hours.
\end{enumerate}

\section{Backgrounds \& Motivation}
\subsection{Learning Embeddings: Transition from Multi-Stage to Single-Stage Training, Integration of Synthetic Data}
The research on learning good embeddings has undergone a significant evolution, shifting from multi-stage to single-stage training methodologies, and from reliance on human-labeled data to the integration of synthetic data. Traditionally, methods including Contriever \cite{izacard2021unsupervised}, OpenAI Embeddings \cite{neelakantan2022text}, E5 \cite{wang2022text}, BGE \cite{xiao2023c}, and Gecko \cite{lee2024gecko} have adopted a multi-stage training paradigm to mitigate the limitations of labeled data in terms of task diversity and language coverage. These approaches involve (1) initial pre-training phase on large-scale, weakly-supervised text pairs using contrastive loss, followed by (2) fine-tuning on smaller, high-quality datasets to enhance the performance.

Recent advancements, however, have seen a move towards single-stage training approaches in models like E5-Mistral \cite{wang2023improving} and GritLM \cite{muennighoff2024generative}. E5-Mistral, in particular, has demonstrated that extensive auto-regressive pre-training enables large language models (LLMs) to acquire robust text representations with minimal fine-tuning needed to transform them into effective embedding models. This finding suggests that contrastive pre-training has a negligible impact on the model quality, indicating that long periods of such pre-training are no longer necessary. This shift highlights a significant trend in the research on embeddings, emphasizing the growing efficacy of synthetic data and streamlined training processes.

Models including E5-Mistral/GritLM, and Gecko generate and utilize well-curated synthetic data for training. They define multiple task categories and use these to generate new query-positive-negative triplets. The former group generates all components of the query-positive-negative triplet, while the latter, exemplified by Gecko, generates only the query and task description, subsequently performing LLM-based positive and negative mining to complete the triplet.

\subsection{The Importance of Data Quality in Training Embedding Models}
The significance of data quality in training embedding models has been extensively emphasized. 

\textbf{SFR} \cite{SFRAIResearch2024} : This research demonstrates that eliminating false negatives, the number of hard negatives, and the impact of hard negative mining significantly affect model performance. It shows that the quality of hard negatives greatly influences the performance, indicating that high-quality hard negatives are crucial for achieving optimal results. 

\textbf{Gecko} \cite{lee2024gecko} : Gecko introduces an LLM-based positive and negative mining method, leveraging large language models (LLMs) to identify more relevant positive passages and suitable hard negatives for generated queries. For a given query and task description, candidate passages are extracted from a corpus pool using a pre-trained embedding model. These candidates are then ranked by an LLM, with the top-ranked passage selected as the positive and the 20th neighbor as the hard negative. Although queries are generated from given passages, this method can change the global positive in approximately 15\% of cases. The performance varies significantly based on the quality of the selected positives and hard negatives, underscoring the importance of these elements.

\textbf{GritLM} \cite{muennighoff2024generative} : In comparisons of training datasets, GritLM highlights that models trained with E5 (synthetic data) outperform those trained with MEDI and MEDI2 (which have better negatives than MEDI) by a wide margin. This superior performance is attributed to the high quality of hard negatives and the diversity of tasks generated by GPT-4 in the E5 dataset. An inspection of samples supports this conclusion, emphasizing the critical role of data quality in the training process.

\subsection{Our Observation: Issues in Data Generation Using GPT-4-Turbo and E5-Mistral's Synthetic Data Strategy}

Using GPT-4-Turbo (gpt-4-turbo-2024-04-09), we generated data by closely adhering to the synthetic data strategy employed by E5-Mistral \cite{wang2023improving}. This strategy includes six tasks: short-long (retrieval), long-short (classification), long-long (matching), short-short (matching), STS (semantic textual similarity), and bitext (translation). For each of these tasks, we categorized the issue types as outlined in Table \ref{tab:category} and identified several data quality issues, which are detailed in Table \ref{tab:sts}, \ref{tab:long-short}, \ref{tab:short-long}, \ref{tab:short-short}, \ref{tab:long-long}, \ref{tab:bitext}.

\begin{table}[!t]
\centering
\caption{Categorized Issue Types for Each Task} \label{tab:category}
\includegraphics[width=1\linewidth]{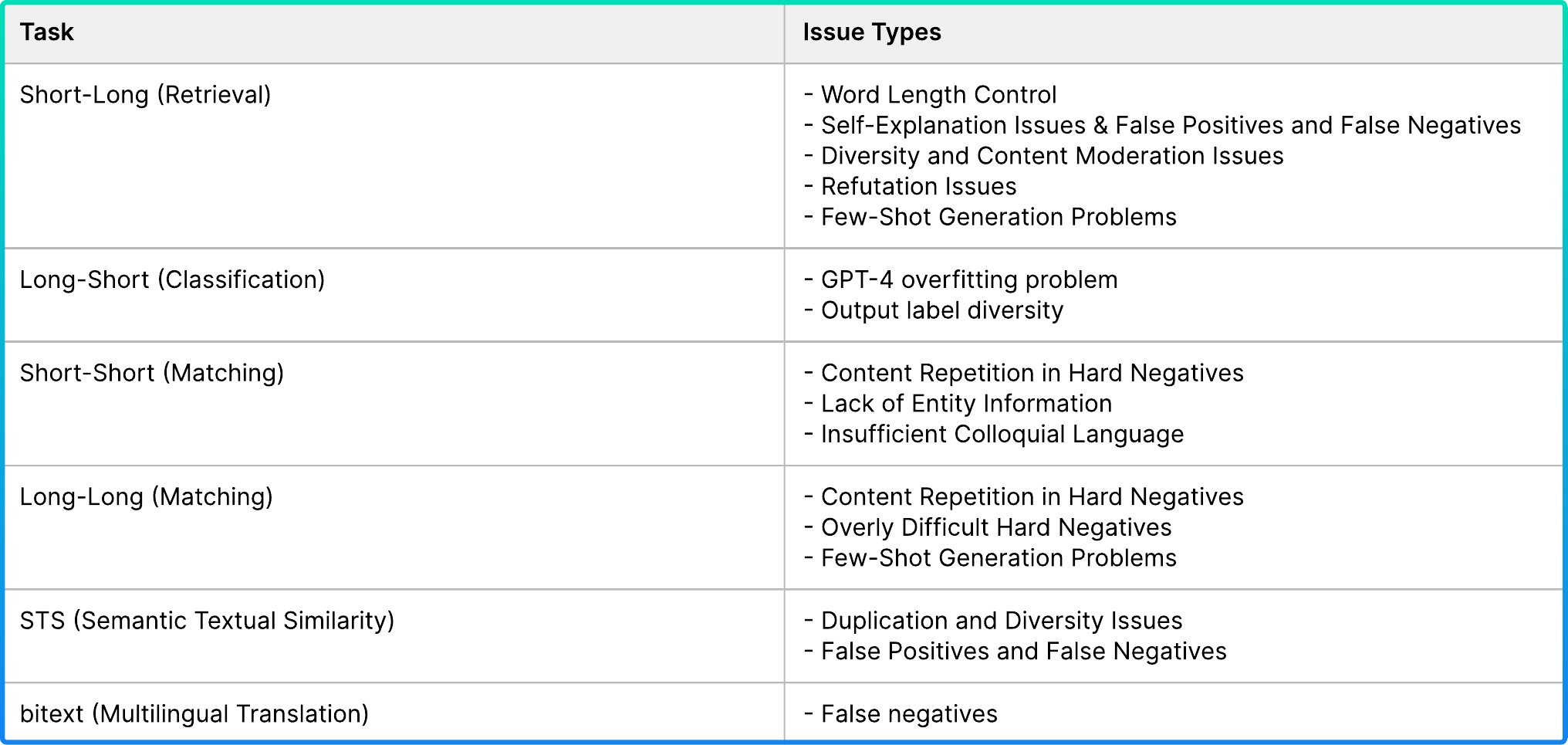}
\end{table}

\begin{table*}[!htb]
\caption{Description of issue types in the STS (Semantic Textual Similarity) task.}
\label{tab:sts}
\centering
\scriptsize
\begin{tcolorbox}
\vspace{-1mm}
{
\textbf{STS (Semantic Textual Similarity)}
\begin{itemize}
    \setlength{\itemindent}{-2em}
    \item Duplication
    \setlength{\itemindent}{-1em}
    \item The tendency to generate identical sentences necessitates deduplication.
    \setlength{\itemindent}{-2em}
    \item Topic Diversity
    \setlength{\itemindent}{-1em}
    \item A high number of similar queries are generated (e.g., quantum computing, data analysis, parks, dogs and cats, "The quick brown fox...").
    \setlength{\itemindent}{-2em}
    \item False Positives
    \setlength{\itemindent}{-1em}
    \item Some generated positives are of lower quality than the generated hard negatives.
    \setlength{\itemindent}{-2em}
    \item False Negatives
    \setlength{\itemindent}{-1em}
    \item In score-based generation for STS tasks, false negatives can occur when the positive and negative examples differ by only 0.5 points in their relatedness scores.
    \setlength{\itemindent}{-2em}
    \item Limited Negative Patterns
    \setlength{\itemindent}{-1em}
    \item General statements: Some negative examples use abstract level of terminologies, e.g., hypernym or holonym.  (e.g., “Multivariable calculus is pivotal in understanding dynamic economic models.” vs. “Calculus is used in some basic economic calculations.”)
    \setlength{\itemindent}{-1em}
    \item Counterfactual statements: Statements that swap subjects and objects (e.g., "He likes ball" vs. "She likes ball").
    \setlength{\itemindent}{-1em}
    \item Lack of Negation: GPT-generated data rarely includes negation patterns (e.g., "I like you" vs. "I don’t like you").
\end{itemize}
}
\vspace{-1mm}
\end{tcolorbox}
\vspace{-3mm}
\end{table*}

\begin{table*}[!htb]
\caption{Description of issue types in the Long-Short (Classification) task.}
\label{tab:long-short}
\centering
\scriptsize
\begin{tcolorbox}
\vspace{-1mm}
{
\textbf{Long-Short (Classification)}
\begin{itemize}
    \setlength{\itemindent}{-2em}
    \item GPT-4 Overfitting Problem
    \setlength{\itemindent}{-1em}
    \item GPT-4-turbo tends to generate specific words with high probability when certain keywords are included, despite using random seed and temperature settings of 1.0.
    \setlength{\itemindent}{-2em}
    \item Output Label Diversity
    \setlength{\itemindent}{-1em}
    \item The typical label types and relationships used for evaluation might not be sufficiently represented in the generated data due to the variety of tasks, which can range from very detailed to very broad.
\end{itemize}
}
\vspace{-1mm}
\end{tcolorbox}
\vspace{-3mm}
\end{table*}

\begin{table*}[!htb]
\caption{Description of issue types in the Short-Long (Retrieval) task.}
\label{tab:short-long}
\centering
\scriptsize
\begin{tcolorbox}
\vspace{-1mm}
{
\textbf{Short-Long (Retrieval)}
\begin{itemize}
    \setlength{\itemindent}{-2em}
    \item Inconsistent Word Length
    \setlength{\itemindent}{-1em}
    \item GPT-4-turbo struggles with controlling the word length, particularly for non-English languages.
    \setlength{\itemindent}{-2em}
    \item Self-Explanation Issues
    \setlength{\itemindent}{-1em}
    \item Rationale Inclusion: The rationale on the positiveness/negativeness of the GPT-generated passages is added to the generated positive/negative passages.
    \setlength{\itemindent}{-1em}
    \item Query Inclusion: Queries are sometimes included verbatim in the passages.
    \setlength{\itemindent}{-2em}
    \item False Positives and False Negatives
    \setlength{\itemindent}{-1em}
    \item Quality of Hard Negatives: Although the generated hard negatives are less relevant to the given query than the positives, some hard negatives generated by the prompts still provide general or partially answerable responses.
    \item Quality of Given Positives: In some cases, the given positive passage is ambiguous, and a more relevant positive sample exists within the passage pool. This causes issues during training because the model may incorrectly treat the more relevant passage as a negative, while using the ambiguous passage as a positive.
    \item False Negatives: Some GPT-generated hard negatives are actually false negatives, being semantically similar to the positives but incorrectly classified as negatives.
    \item Higher Incidence in Non-English Languages: False positives and false negatives occur more frequently in non-English languages.
    \setlength{\itemindent}{-2em}
    \item Lack of Representation
    \setlength{\itemindent}{-1em}
    \item Synthetic data lacks representation on topics such as same-sex marriage, feminism, and abortion.
    \setlength{\itemindent}{-2em}
    \item Safety Concerns
    \setlength{\itemindent}{-1em}
    \item Socially sensitive topics such as climate change and gun ownership are affected by content moderation, restricting discussion or representation.
    \setlength{\itemindent}{-2em}
    \item Incorrect Refutations as False Positives
    \setlength{\itemindent}{-1em}
    \item When generating refutations, the model sometimes produces supportive results instead, resulting in false positives.

\end{itemize}
}
\vspace{-1mm}
\end{tcolorbox}
\vspace{-5mm}
\end{table*}

\begin{table*}[!htb]
\caption{Description of issue types in the Short-Short (Matching) task.}
\label{tab:short-short}
\centering
\scriptsize
\begin{tcolorbox}
\vspace{-1mm}
{
\textbf{Short-Short (Matching)}
\begin{itemize}
    \setlength{\itemindent}{-2em}
    \item Content Repetition in Hard Negatives
    \setlength{\itemindent}{-1em}
    \item Some hard negatives repeat the content of the positives verbatim, making it challenging to classify them as true negatives.
    \setlength{\itemindent}{-2em}
    \item Lack of Entity Information
    \setlength{\itemindent}{-1em}
    \item The generated data lacks sufficient information about entities such as personal names, place names, movie titles, and game titles.
    \setlength{\itemindent}{-2em}
    \item Insufficient Colloquial Language
    \setlength{\itemindent}{-1em}
    \item There is a lack of colloquial language in the generated data, which is necessary for creating more realistic and relatable passages.
\end{itemize}
}
\vspace{-3mm}
\end{tcolorbox}
\vspace{-7mm}
\end{table*}

\begin{table*}[!h]
\caption{Description of issue types in the Long-Long (Matching) task.}
\label{tab:long-long}
\centering
\scriptsize
\begin{tcolorbox}
\vspace{-1mm}
{
\textbf{Long-Long (Matching)}
\begin{itemize}
    \setlength{\itemindent}{-2em}
    \item Content Repetition in Hard Negatives
    \setlength{\itemindent}{-1em}
    \item Some hard negatives repeat the content of the positives verbatim, making it challenging to classify them as true negatives.
    \setlength{\itemindent}{-2em}
    \item Overly Challenging Hard Negatives
    \setlength{\itemindent}{-1em}
    \item The generated hard negatives are often too difficult, leading to a lack of clearly distinguishable negatives.
\end{itemize}
}
\vspace{-1mm}
\end{tcolorbox}
\vspace{-3mm}
\end{table*}

\begin{table*}[!htb]
\caption{Description of issue types in the Bitext (Multilingual Translation) task.}
\label{tab:bitext}
\centering
\scriptsize
\begin{tcolorbox}
\vspace{-1mm}
{
\textbf{Bitext (Multilingual Translation)}
\begin{itemize}
    \setlength{\itemindent}{-2em}
    \item High Incidence of False Negatives
    \setlength{\itemindent}{-1em}
    \item A significant number of false negatives are produced, regardless of score differences (between positives and negatives) as defined in the E5-Mistral prompt. To be specific, many negatives have the same meaning as the positives, with only slight difference in wording. This is because, in the multilingual settings, the generated negative is often just another translated version of the query. This issue is pronounced in multi-lingual settings, where hard negatives in translation tasks can inadvertently become positives.
\end{itemize}
}
\vspace{-1mm}
\end{tcolorbox}
\vspace{-3mm}
\end{table*}

\section{Training Dataset}

Recall that our focus is to improve the text retrieval using advanced data refinement methods, to both (1) existing benchmark dataset, and (2) highly tailored synthetic dataset generated via LLMs. In this section, we provide how our training data differs from the dataset used in conventional high-performing embedding models. To be specific, the training data used for models originated from E5-Mistral can be summarized as below:

\begin{itemize}
    \item \blue{E5-Mistral} : E5 data (human labeled) + synthetic data
    \item \orange{SFR}: a specially curated dataset comprising MTEB (human labeled)
    \item \green{GritLM}: E5S data (human labeled) + synthetic data
    \item \red{Ours}: E5S data (human labeled) + synthetic data (well-curated)
\end{itemize}

As demonstrated in Table \ref{tab: trainingdataset}, the SFR model is trained on a specially curated dataset of MTEB tasks, including Retrieval, Clustering, Classification, STS, and Reranking tasks, while excluding development and testing sets. Notably, SFR uniquely treats the labels in clustering and classification tasks as documents, applying contrastive loss exclusively to their respective negatives and omitting in-batch negatives. This method utilizes the \textbf{labels in the evaluation tasks directly during training}, which SFR hypothesizes encourages models to regularize embeddings based on high-level concepts, resulting in better separation of data across different domains. Instead, we focused exclusively on the training dataset used by E5-Mistral. Specifically, we utilized the same dataset composition as in the E5S data from GritLM, augmenting it with S2ORC data to enhance the training dataset used by E5-Mistral. We further analyzed and refined the dataset with the task list used in the E5S data.

Recall that our objective is to \textbf{produce high-quality synthetic data using GPT and verify its effectiveness}. Our synthetic data does not incorporate labels but follows the conventional triplet form of query-positive-negative, comprised solely of documents. By doing so, we seek to reinforce the human-labeled benchmark dataset with high-quality synthetic data, ensuring that the synthetic data never sees the labels in the evaluation tasks directly. This method adheres to E5-Mistral data structures while leveraging the advanced capabilities of GPT to enhance data quality.

All in all, our approach seeks to improve the quality and reliability of LLM-generated data by focusing on sophisticated synthetic data generation. This method stands in contrast to SFR's label-utilizing strategy, proposing that a well-crafted synthetic dataset can effectively support model training and generalization without direct reliance on evaluation task labels. The impact of adding the data used by SFR to our dataset setting is remained as a future work.

\begin{table}[!hb]
\centering
\caption{Comparison of the benchmark dataset used for training various models. Considering the fact that MTEB benchmark is used to test each embedding model, it is less desired to train with datasets included in MTEB. However, one high-performing baseline (SFR model) is trained on a specially curated dataset comprising MTEB tasks, including Retrieval, Clustering, Classification, STS, and Reranking tasks, while excluding development and testing sets. Instead, the dataset composition of our benchmark dataset is identical to that of E5S data used in GritLM \cite{muennighoff2024generative}, which is the concatenation of the training dataset used by E5-Mistral and S2ORC data. Here, S2ORC may overlap with the Arxiv-related tasks in MTEB.} \label{tab: trainingdataset}
\includegraphics[width=0.95\linewidth]{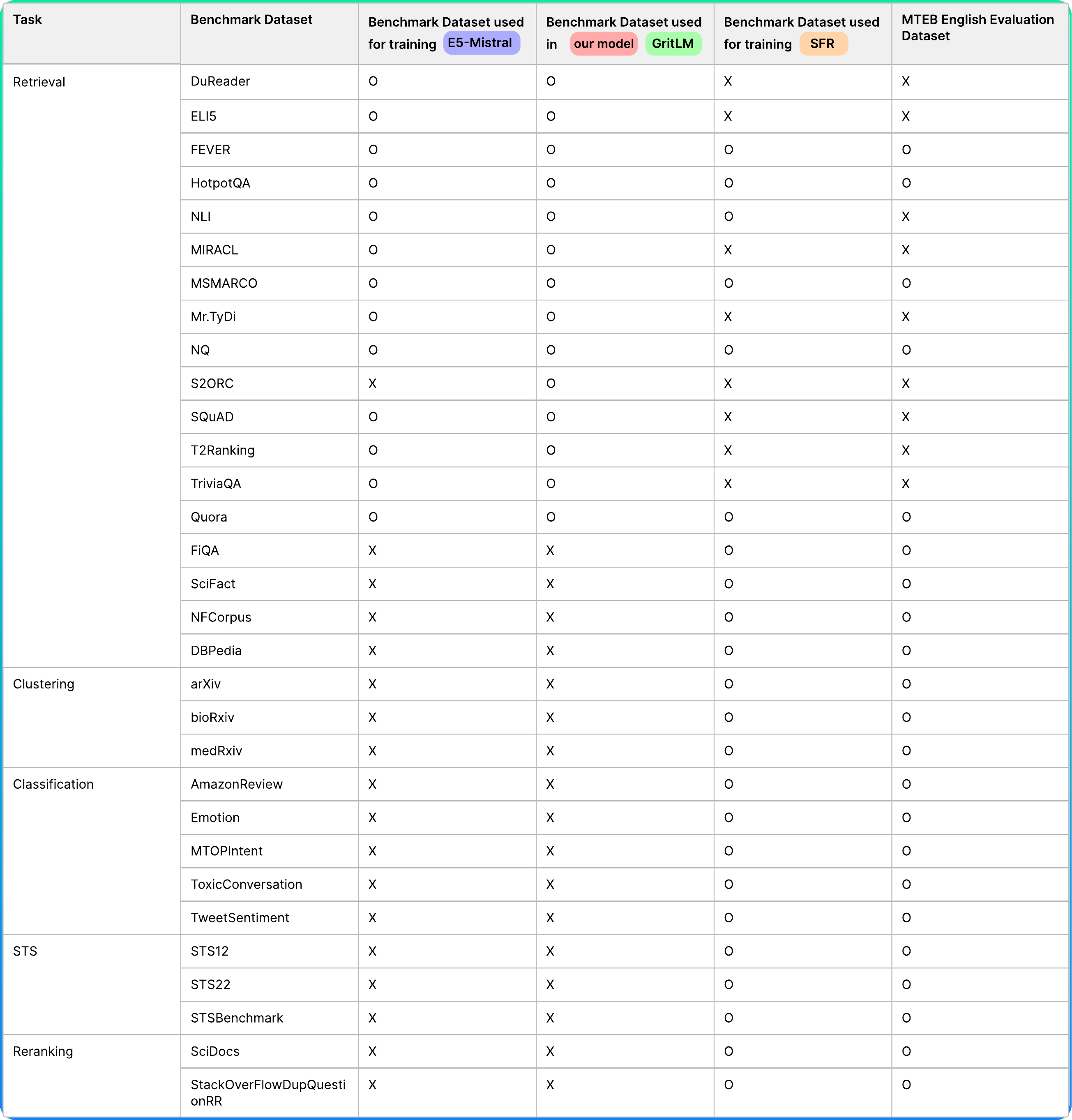}
\end{table}

\section{Proposed Data Refinement Methods}
In this section, we provide details of our proposed data refinement methods used on (1) Benchmark Dataset and (2) Synthetic Data. Our methods include sophisticated data crafting, data filtering, and negative mining techniques.

\begin{figure}[!htb]
\centering
\includegraphics[width=1\linewidth]{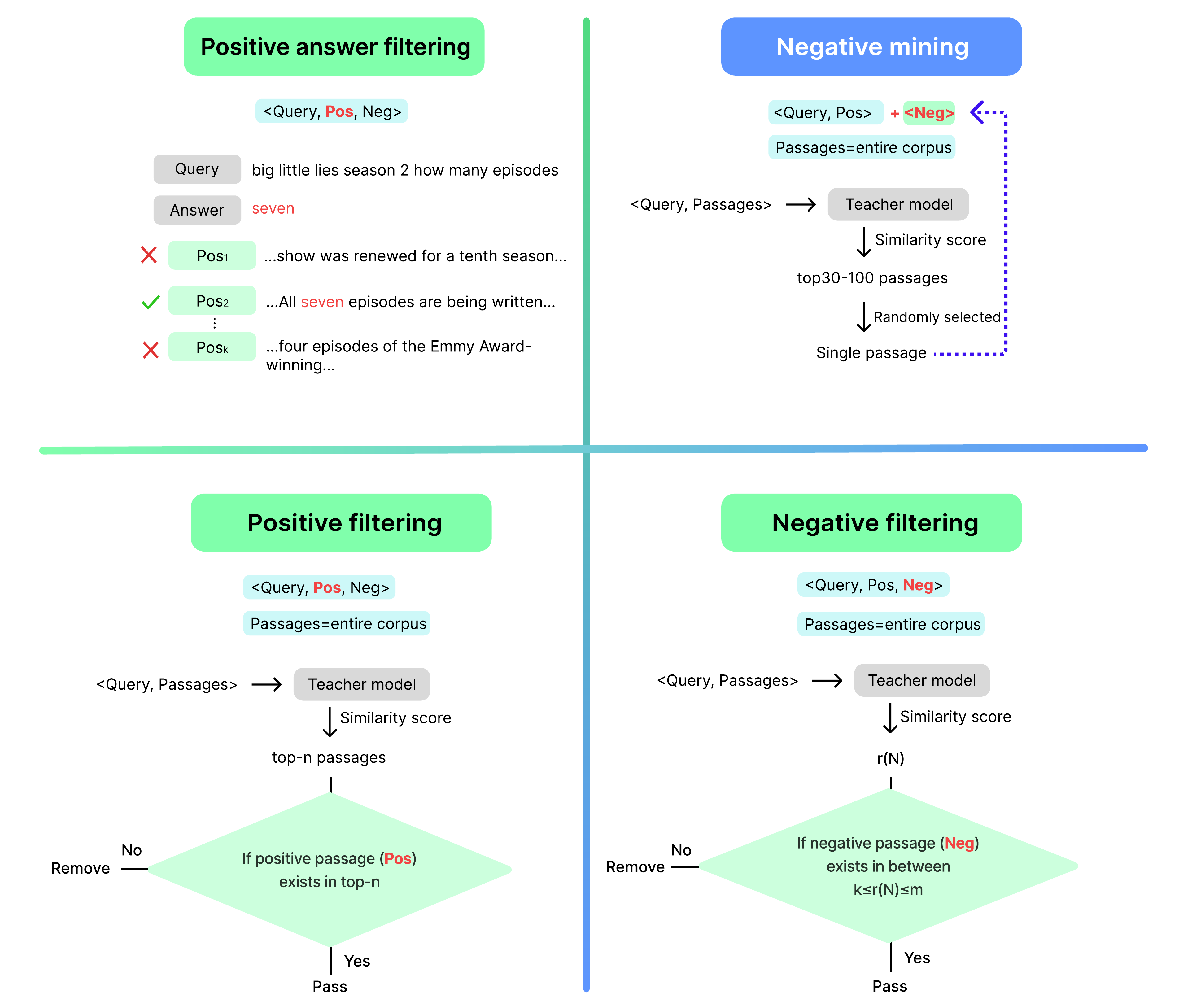}
\caption{Overview of our proposed methods of refining the Benchmark Dataset.}
\label{fig:refinement}
\end{figure}

\subsection{Data Refinement on Benchmark Dataset}
For each task, we applied various methodologies to identify the most effective combination, selectively implementing the best-performing strategies for each task. These methods include:
\begin{itemize}
    \item \textbf{Data Source Selection}: When multiple data sources are available (e.g., HotpotQA from KILT or DPR), we tested each source and selected the most suitable one.
    \item \textbf{Positive Answer Filtering}: Utilizing only passages that contain the answer.
    \item \textbf{Positive Filtering}: Incorporating positives only if they are within the top-n rankings of the teacher model.
    \item \textbf{Negative Mining}: Using samples within the top 30-100 rankings of the teacher model as negatives.
    \item \textbf{Negative Filtering}: Implementing various strategies using teacher models, where negatives $\left(N\right)$ are considered only if their rankings $\left(r\right)$ fall within a specific top $\left(n\right)$ to $\left(m\right)$ range of the teacher model's rankings, i.e., $\left(k \leq r(N) \leq m\right)$.
\end{itemize}

The performance varies significantly depending on the combination of methods used for each task. For instance, in some tasks where answers are present, filtering positives based on the inclusion of the answer may enhance the performance, while in other tasks, this approach may not be beneficial. It is crucial to tailor the refinement strategies to the specific requirements and characteristics of each task to achieve optimal performance. The specific processing methods applied for each task can be reviewed in Figure \ref{fig:refinement}. 

\begin{table}[!htb]
\caption{Refinement methods for enhancing the quality of Synthetic Data across six tasks, demonstrating up to two independent methods per task.} \label{tab:refinement-synthetic}
\label{tag:refinement}
\centering
\includegraphics[width=1\linewidth]{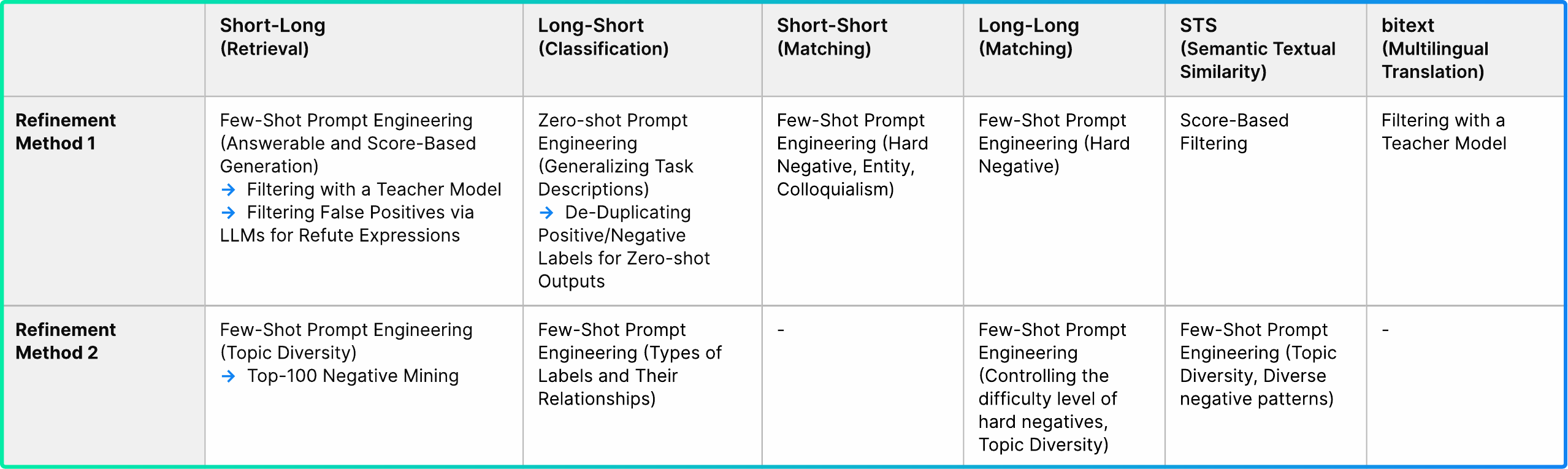}
\end{table}

\subsection{Data Refinement on Synthetic Data}
As detailed in Table \ref{tab:refinement-synthetic}, the quality of synthetic data is scrutinized for each task, employing various types of few-shot prompt engineering, filtering, and negative mining to improve overall data quality. In Table \ref{tab:huge}, we provide detailed description of each issue and our proposed solution to handle the issue.

\section{Training Details: Mostly Follow E5-Mistral and SFR}
Our experimental setup integrates several strategies and techniques, drawing from SFR and E5-Mistral, aiming to optimize the model performance and the training efficiency.
\paragraph{Contrastive Loss} We utilize the standard InfoNCE loss over in-batch negatives and hard negatives, formulated as follows:
\begin{center}
    $\min \mathcal{L} = -\log \frac{\phi(q, d)}{\phi(q, d) + \sum_{n_i \in \mathcal{N}} \phi(q, n_i)}$,
\end{center}
where $\mathcal{N}$ denotes the set of all negative documents, and $\phi(q, d)$ is a function that computes the matching score between query $q$ and document $d$. Following E5-Mistral, we adopt the temperature-scaled cosine similarity function:
\begin{center}
    $\phi(q, d) = \exp\left(\frac{1}{\tau} \cos(h_q, h_d)\right)$,
\end{center}
where $h_q$ and $h_d$ are the embeddings of the query and the document, respectively, and the temperature is set to $\tau =0.02$.

\paragraph{Fine-Tuning Procedure} We fine-tuned the E5-Mistral-7b-instruct model\footnote{\url{https://huggingface.co/intfloat/E5-Mistral-7b-instruct?ref=blog.salesforceairesearch.com}} using a batch size of 2,080 and a learning rate of 1e-4, starting with a warm-up phase followed by linear decay. Despite the advantage of larger batch sizes noted by SFR, increasing from 2,048 to 8,192 showed no significant performance change. This fine-tuning process took approximately 30 hours on four A100 GPUs.
\paragraph{Maximum In-Device Batch Size} We used A100 80GB GPUs to train with the maximum in-device batch size, akin to the SRF approach, considering the impact of the number of negative samples.
\paragraph{Sequence Length} While SRF used a maximum sequence length of 128 for queries and 256 for documents, we opt for a maximum sequence length of 4K, aiming to generalize better for (long query, long passage) tasks. For our experiments, we adhere to the settings of E5-Mistral, limiting evaluations to the first 512 tokens for efficiency, despite the model's capability to handle longer sequences.
\paragraph{Task-Homogeneous Batching} According to SFR, constructing batches with samples from a single task enhances the difficulty of in-batch negatives and improves retrieval task performances. We followed such setting.
\paragraph{Number of Hard Negatives} Although SFR found that using seven hard negatives per query yields the best results, our experiments showed diminishing returns with more hard negatives as data quality improved. Therefore, we opt to use only one hard negative, using the mE5-base model as the teacher for negative mining, unlike SFR's use of the BGE-base model\footnote{\url{https://huggingface.co/BAAI/bge-base-en}}.
\paragraph{Top 30-100 Negative Mining} We found that negative sampling significantly affects the performance more than the in-device batch size. As reported by SFR, selecting the top 30-100 negative samples impacts the performance most significantly, underscoring the importance of precise negative sample selection. 
\paragraph{LoRA Adapters} We added LoRA adapters with alpha 32 and rank 16 to all linear layers. Despite no significant performance difference between ranks 16 and 8, we chose rank 16 for its greater stability and reduced fluctuation during training.
\paragraph{E5-Mistral Style One-Sided Instruction Prefix} The E5-Mistral approach uses one-sided instructions for asymmetric datasets, where only queries receive instructions, allowing the document corpus to be encoded once, cached, and reused across tasks. As below, GritLM explains that even symmetric tasks are handled one-sidedly during training but evaluated in a two-sided format, ensuring the consistency and the reliability of similarity measures. 

"During training, even symmetric tasks are handled in a one-sided manner, though they are evaluated in a two-sided format. This is feasible because the cosine similarity function used during training is transitive. Therefore, if a sentence with instructions (A) is similar to a sentence without instructions (B), and B is similar to another sentence with instructions (C), it can be inferred that A is also similar to C. This ensures consistency and reliability in similarity measures despite the one-sided instruction approach during training" \cite{muennighoff2024generative}.

\paragraph{In-Batch Negative Strategy in Clustering and Classification} The embedding model for classification or clustering tasks can be misled by the in-batch negatives technique, as the ``passage'' within a batch might belong to the same class and therefore are not actual negatives. SFR uniquely treats labels as documents in clustering and classification tasks, applying contrastive loss exclusively to their respective negatives and avoiding in-batch negatives to prevent misleading the embedding model. Conversely, Gecko \cite{lee2024gecko} addresses the issue of false negatives by assigning a unique ID to each triple (query, positive example, negative example), making in-batch negatives trivial for the model to distinguish. For long-short and long-long synthetic data tasks, our method applied the unique ID method, similar to Gecko, exclusively in few-shot scenarios where generative diversity decreases, and did not use it in zero-shot scenarios.
\paragraph{Other Training Techniques} We applied various advanced training techniques, including gradient checkpointing, mixed precision training (fp16), and DeepSpeed ZeRO-3, to enhance training efficiency and performance.

\begin{table}[!h]
\caption{Identified Issues Affecting the Quality of Synthetic Data.}
\label{tab:huge_1}
\centering
\includegraphics[width=1\linewidth]{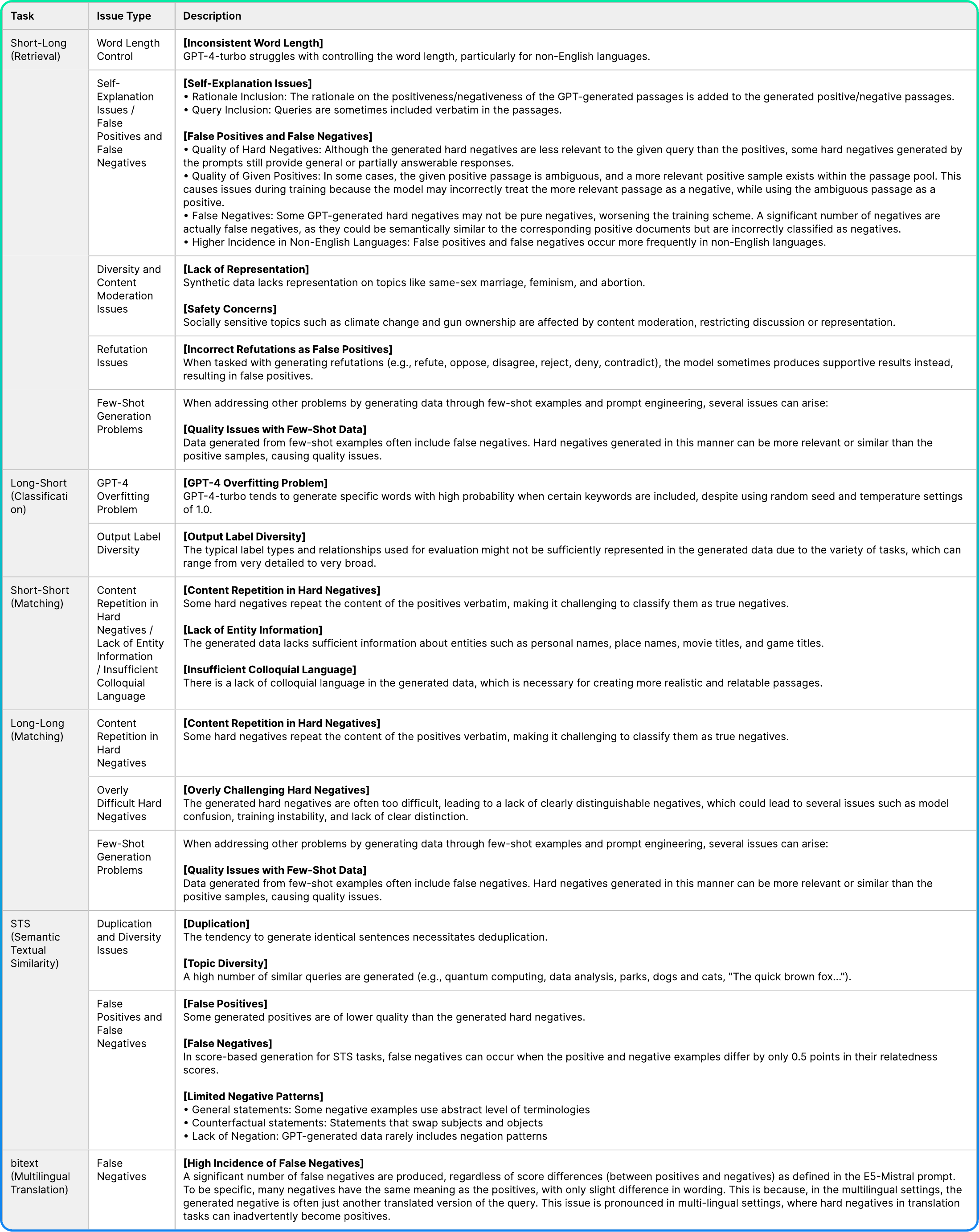}
\end{table}

\begin{table}[!h]
\caption{Summary of Proposed Solutions for Improving Synthetic Data Quality.}
\label{tab:huge_2}
\centering
\includegraphics[width=1\linewidth]{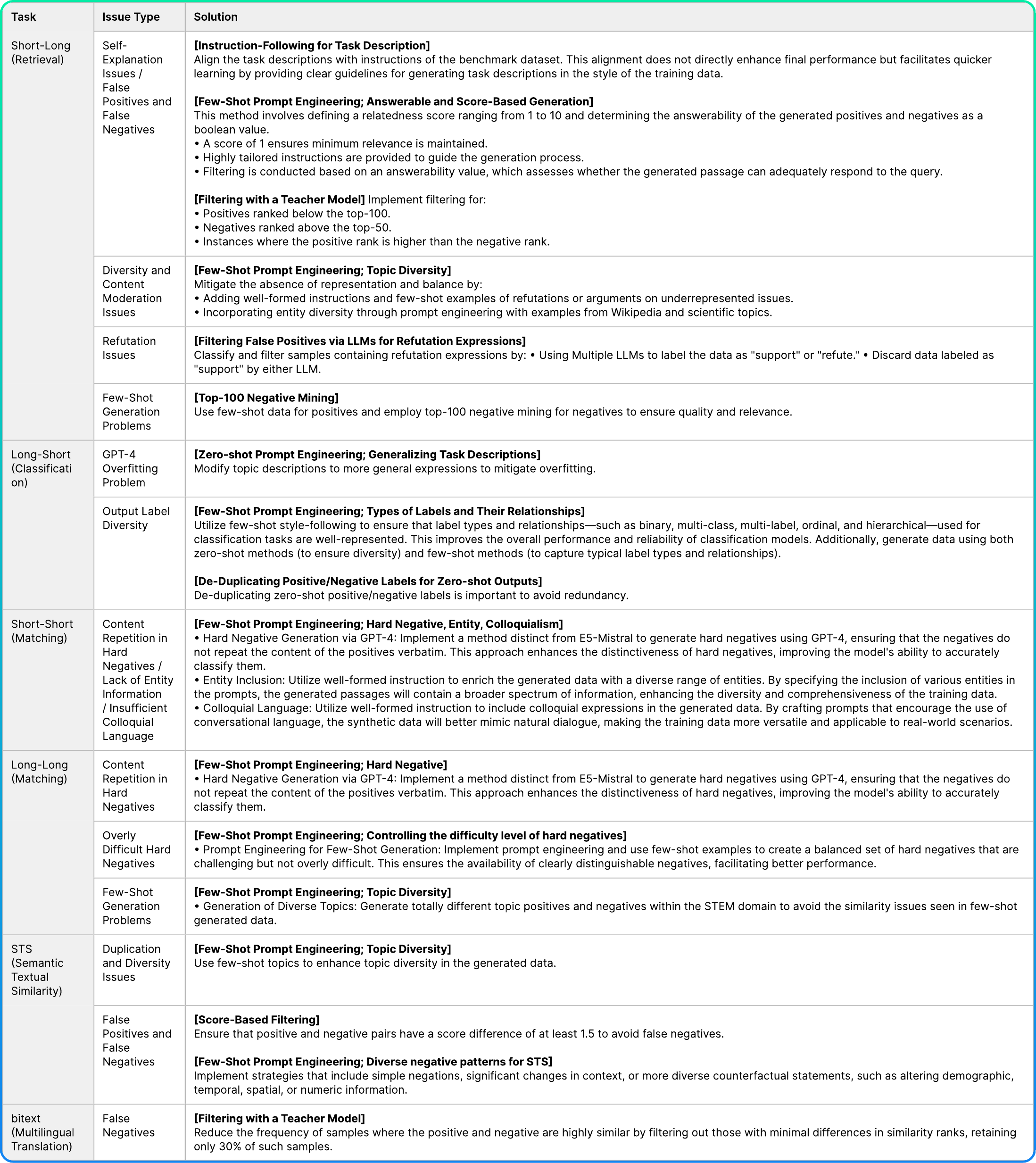}
\end{table}

\section{Homogeneous Task Ordering and Mixed Task Fine-tuning}
In this section, we provide details of our proposed training methods, dubbed as Homogeneous Task Ordering and Mixed Task Fine-tuning , details of which are in the upcoming subsections. We first train a model using Homogeneous Task Ordering method, and then fine-tune this model using Mixed Task Fine-tuning method.

\subsection{Homogeneous Task Ordering}
Recall that existing embedding models (including SFR) rely on task-homogeneous batching, which constructs each batch with the samples from identical task. We follow this way of constructing the batches, but added another constraint on the training, \textbf{which specifies the orders of tasks handled in the consecutive batches}. To be specific, our batches of 1-epoch training is composed of $n$ blocks, where we assigned samples in the batches and blocks in a way that the task order within each block is identical across $n$ blocks. The primary objective of this proposed Homogeneous Task Ordering method is to clearly track how the performance changes as one block of batches (having specific task orders) are loaded. This provides us more precise insights on the effects of task ordering on the model performance.

\subsection{Mixed Task Fine-Tuning}
Although task-homogeneous batching is prevalent method used in existing works, it also has some limitations. This batching strategy limits the learning process to one task per gradient update, leading to potential catastrophic forgetting. Catastrophic forgetting occurs when a neural network loses previously acquired knowledge about one task due to exclusive focus on another, resulting in performance fluctuations observed at specific intervals. Additionally, the randomization of task sequence can delay the training of certain tasks, further aggravating the problem.

Motivated by these issues, we use Mixed Task Fine-tuning, defined as below:
\begin{itemize}
    \item \textbf{Mixed Task Fine-Tuning}: While maintaining a homogeneous batch within the device, this method involves mixing different tasks to compose each batch. This strategy aims to mitigate the risks associated with learning in a homogeneous environment by diversifying the learning inputs.
\end{itemize}
Our training framework involves an initial phase of homogeneous task fine-tuning for one complete epoch, followed by mixed task fine-tuning conducted over a few steps. This structured approach is designed to optimize learning across diverse tasks, thereby reducing performance fluctuations and the impact of catastrophic forgetting on untrained tasks at the current step.

In our experiments, mixed task fine-tuning within 20 steps demonstrated the best performance. Extending the training beyond this point resulted in a decline in performance.

\section{Streamlined Evaluation}
\begin{figure}[tb!]
\centering
\includegraphics[width=0.9\linewidth]{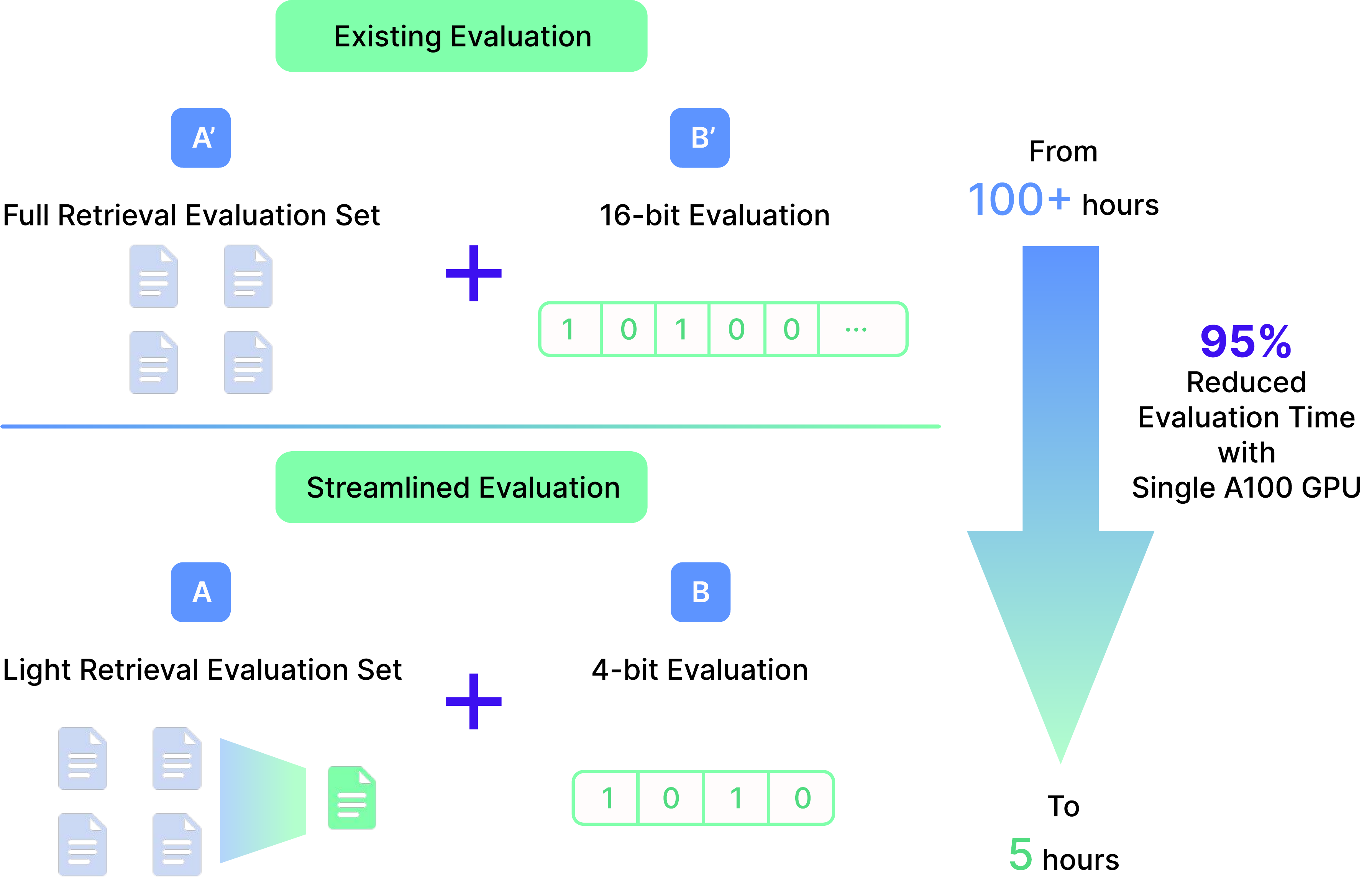}
\caption{Strategies for streamlining the evaluation process.}
\label{fig:eval}
\end{figure}

To facilitate rapid experimentation, we employed a combination of strategies that streamline the evaluation process, as demonstrated in Figure \ref{fig:eval}. These strategies include (a) the use of a light retrieval evaluation set and (b) evaluations conducted using 4-bit precision. Collectively, these approaches enable quick and effective experimentation. Using a single GPU, one checkpoint can be evaluated in approximately 5 hours, with retrieval tasks specifically taking about 4 hours.

\subsection{Light Evaluation Set} 
\paragraph{For Retrieval Task} We observed that evaluating the performance of a model on Retrieval tasks takes much longer than the evaluation on other tasks in MTEB benchmarks. Thus, in order to check the performance of a model rapidly, we designed a light retrieval evaluation set, which has negligible performance differences compared to the performances observed for the full evaluation set. This approach allows efficient assessment of the model's performance on retrieval tasks without significantly compromising accuracy, thus optimizing the overall evaluation process.
\begin{itemize}
    \item \textbf{Corpus Sampling}: Among the entire corpus, the top 50 most relevant corpora for each query are extracted via the previously trained teacher model. Subsequently, sets will be created to eliminate overlaps, thereby reducing the number of corpora utilized for evaluation.
    \item \textbf{Query Sampling}: We sample 20\% of queries with the balanced labels to further streamline the process.
\end{itemize}

\paragraph{For Other Tasks (Classification, Clustering, Pair Classification, Re-ranking, STS, Summarization)} We omitted evaluation on large-scale sub-tasks.

\subsection{4-bit Evaluation}
Using 4-bit precision for evaluations allow a single GPU to process more samples compared to the case of using 16-bit precision. This results in a substantial increase in the processing speed, with observed improvements of up to approximately 40\% in our GPU setting. This significant speedup comes without the cost of accuracy.

\begin{table}[tb!]
\centering
\caption{Comparison with publicly accessible models.}\label{tab:oss}
\includegraphics[width=1\linewidth]{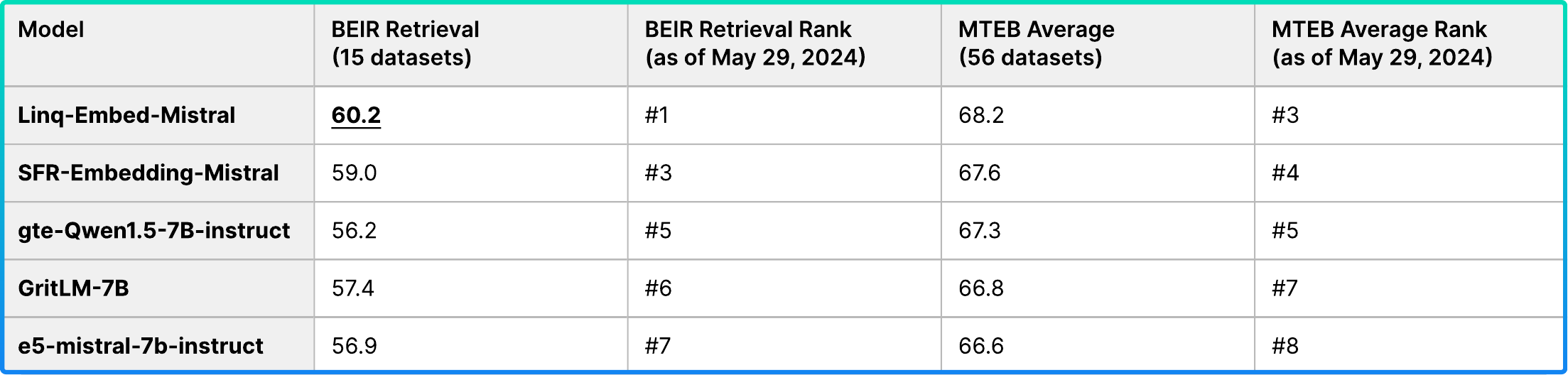}
\end{table}

\begin{table}[t!]
\centering
\caption{Comparison with commercial models.}\label{tab:commercial}
\includegraphics[width=1\linewidth]{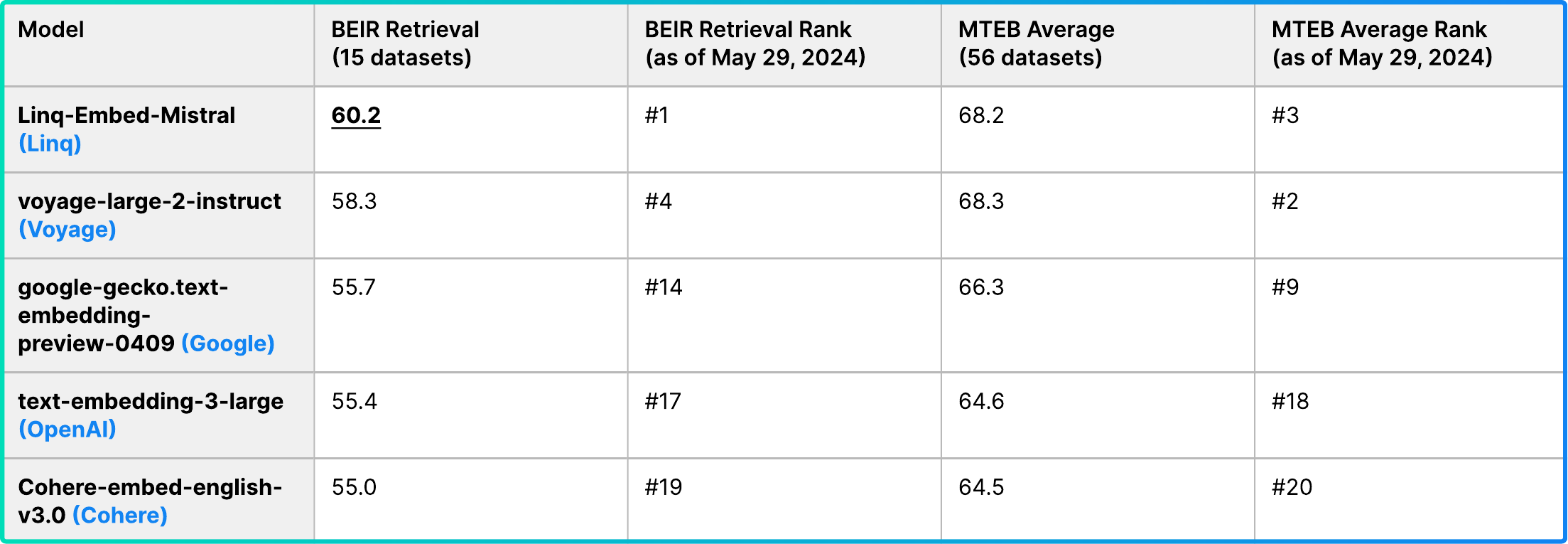}
\end{table}

\newpage
\section{Full Evaluation on MTEB}
The Massive Text Embedding Benchmark (MTEB) stands as the most comprehensive benchmark for evaluating embedding models, incorporating 56 datasets across seven task types: classification, clustering, pair classification, re-ranking, retrieval, STS, and summarization.

The key experimental points are:
\begin{itemize}
    \item Linq-Embed-Mistral performs well in the MTEB benchmarks, with an average score of 68.2 across 56 datasets. This places it 1st among publicly accessible models listed on the MTEB leaderboard and 3rd overall.
    \item The model shows a significant enhancement in the retrieval performance, ranking 1st among all models listed on the MTEB leaderboard with a performance score of 60.2.
    \begin{itemize}
        \item Within the Mistral Model Series, a suite of models based on the foundational Mistral architecture, SFR enhances E5-Mistral by adding a specially curated dataset of MTEB tasks. In contrast, our approach focuses solely on \textbf{creating and integrating more sophisticated synthetic datasets}. This has increased our model's score from 56.9 for E5-Mistral and 59.0 for SFR to an 60.2.
    \end{itemize}
\end{itemize}

\begin{table}[!h]
\centering
\caption{Comparison with the models that tops the MTEB leaderboard (as of May 29, 2024), highlighting the first and second place items in each task using bold and underlined formatting.}\label{tab:leaderboard}
\includegraphics[width=1\linewidth]{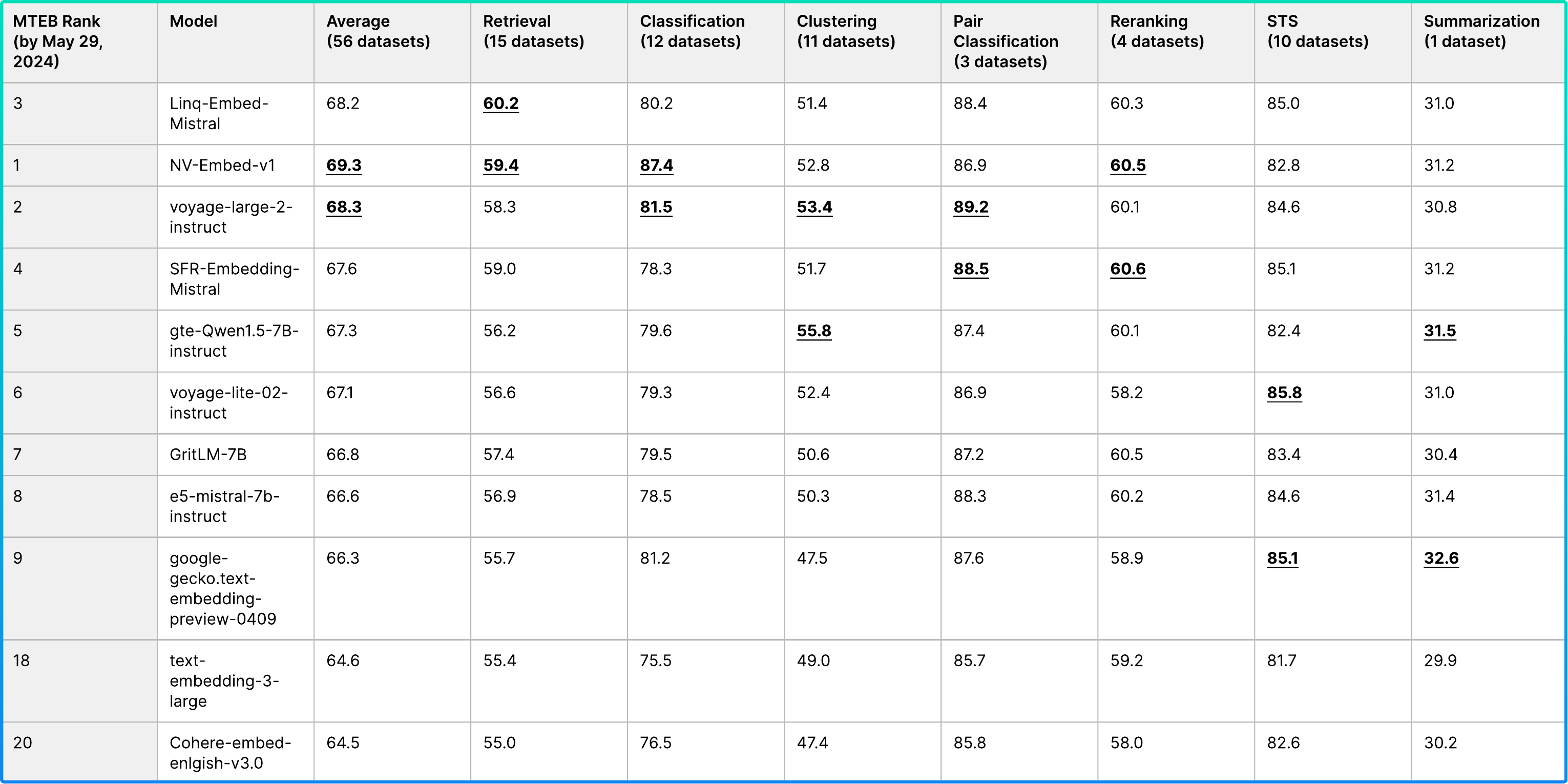}
\end{table}
\begin{ack}
    Contributions to the development of Linq-Embed-Mistral were made by
    
    \href{mailto:soulloan@gmail.com}{\textbf{Junseong Kim}}  (Project Leader; GPT Data Strategy, Experiment Design, Technical Guidance and Advice) \href{mailto:seolhwa.lee.august@gmail.com}{\textbf{Seolhwa Lee}} (AI Researcher; Dataset Filtering \& Mining, Modeling Experiments, GPT Data Generation) \href{mailto:kog0712@snu.ac.kr}{\textbf{Jihoon Kwon}} (AI Intern; Benchmark Datasets for Training, Dataset Filtering \& Mining, Data Training Pipelines, Modeling Experiments, MTEB Evaluation) \href{mailto:gsm1014@snu.ac.kr}{\textbf{Sangmo Gu}} (AI Intern; GPT Data Strategy, GPT Data Filtering, GPT Data Generation) \href{mailto:yjkim.stat@yonsei.ac.kr}{\textbf{Yejin Kim}} (AI Intern; Baseline Models) \href{mailto:kveldsstjerne@snu.ac.kr}{\textbf{Minkyung Cho}} (AI Intern; Baseline GPT Data) \href{mailto:jysohn1108@gmail.com}{\textbf{Jy-yong Sohn}} (Advisor; Technical Guidance and Advice) \href{mailto:jacob.choi@getlinq.com}{\textbf{Chanyeol Choi}} (Advisor; Technical Guidance and Advice).
\end{ack}


\bibliography{reference}
\bibliographystyle{ieeetr}





\end{document}